\documentclass[11pt,a4paper]{article}
\usepackage[hyperref]{emnlp2020}
\usepackage{times}
\usepackage{latexsym}

\usepackage[utf8]{inputenc}
\usepackage{xcolor}
\usepackage{booktabs}

\usepackage{graphicx}

\usepackage{subcaption}

\usepackage{microtype}

\usepackage{enumitem}  %
\setlist[itemize]{leftmargin=*}
\setlist[enumerate]{leftmargin=*}

\usepackage[frozencache]{minted}
\definecolor{bggray}{rgb}{0.1,0.1,0.1}

\usepackage{listings}
\definecolor{codegreen}{rgb}{0,0.6,0}
\definecolor{codegray}{rgb}{0.5,0.5,0.5}
\definecolor{codepurple}{rgb}{0.58,0,0.82}
\definecolor{backcolour}{rgb}{0.95,0.95,0.92}

\usepackage{hyperref}
\usepackage{cleveref}

\usepackage{outlines}   %
\usepackage{todonotes}  %
\usepackage[normalem]{ulem}  %

\aclfinalcopy %

\title{The Language Interpretability Tool:\\ Extensible, Interactive Visualizations and Analysis for NLP Models}

\author{
  Ian Tenney,\Thanks{~~Equal contribution.}\textsuperscript{ } James Wexler,\footnotemark[1]\textsuperscript{ } Jasmijn Bastings, Tolga Bolukbasi,\\ 
  \textbf{Andy Coenen, Sebastian Gehrmann, Ellen Jiang, Mahima Pushkarna,} \\
  \textbf{Carey Radebaugh, Emily Reif, Ann Yuan} \\
  Google Research \\
  \texttt{\{iftenney,jwexler\}@google.com} \\
}

\date{}

\begin{document}

\maketitle
\begin{abstract}
We present the Language Interpretability Tool (LIT), an open-source platform for visualization and understanding of NLP models. We focus on core questions about model behavior: Why did my model make this prediction? When does it perform poorly? What happens under a controlled change in the input? LIT integrates local explanations, aggregate analysis, and counterfactual generation into a streamlined, browser-based interface to enable rapid exploration and error analysis. We include case studies for a diverse set of workflows, including exploring counterfactuals for sentiment analysis, measuring gender bias in coreference systems, and exploring local behavior in text generation. LIT supports a wide range of models---including classification, seq2seq, and structured prediction---and is highly extensible through a declarative, framework-agnostic API. LIT is under active development, with code and full documentation available at \url{https://github.com/pair-code/lit}.\footnote{A video walkthrough is available at \url{https://youtu.be/j0OfBWFUqIE}.}

\end{abstract}

\section{Introduction}
\label{sec:intro}

Advances in modeling have brought unprecedented performance on many NLP tasks \citep[e.g.][]{Wang_GLUE}, but many questions remain about the behavior of these models under
domain shift \citep{blitzer2007domain} and adversarial settings \citep{jia-liang-2017-adversarial}, and for their tendencies to behave according to social biases \citep{bolukbasi2016man,caliskan2017semantics} or shallow heuristics \citep[e.g.][]{mccoy-etal-2019-right,poliak2018hypothesis}.
For any new model, one might want to know: What kind of examples does my model perform poorly on? Why did my model make this prediction? And critically, does my model behave consistently if I change things like textual style, verb tense, or pronoun gender? Despite the recent explosion of work on model understanding and evaluation \citep[e.g.][]{belinkov-etal-2020-interpretability,blackbox2019,ribeiro2020checklist},
there is no ``silver bullet'' for analysis. Practitioners must often experiment with many techniques, looking at local explanations, aggregate metrics, and counterfactual variations of the input to build a full understanding of model behavior.

Existing tools can assist with this process, but many come with limitations: offline tools such as TFMA \citep{mewald2018tfma} can provide only aggregate metrics, interactive frontends \citep[e.g.][]{wallace-etal-2019-allennlp} may focus on single-datapoint explanation, and more integrated tools \citep[e.g.][]{what-if-tool,mothilal2020explaining,strobelt2018seq2seqvis} often work with only a narrow range of models. Switching between tools or adapting a new method from research code can take days of work, distracting from the real task of error analysis.
An ideal workflow would be seamless and interactive: users should see the data, what the model does with it, and why, so they can quickly test hypotheses and build understanding.

\begin{figure*}[t!]
    \centering
    \includegraphics[width=\textwidth,clip,trim=0mm 17mm 0mm 0mm]{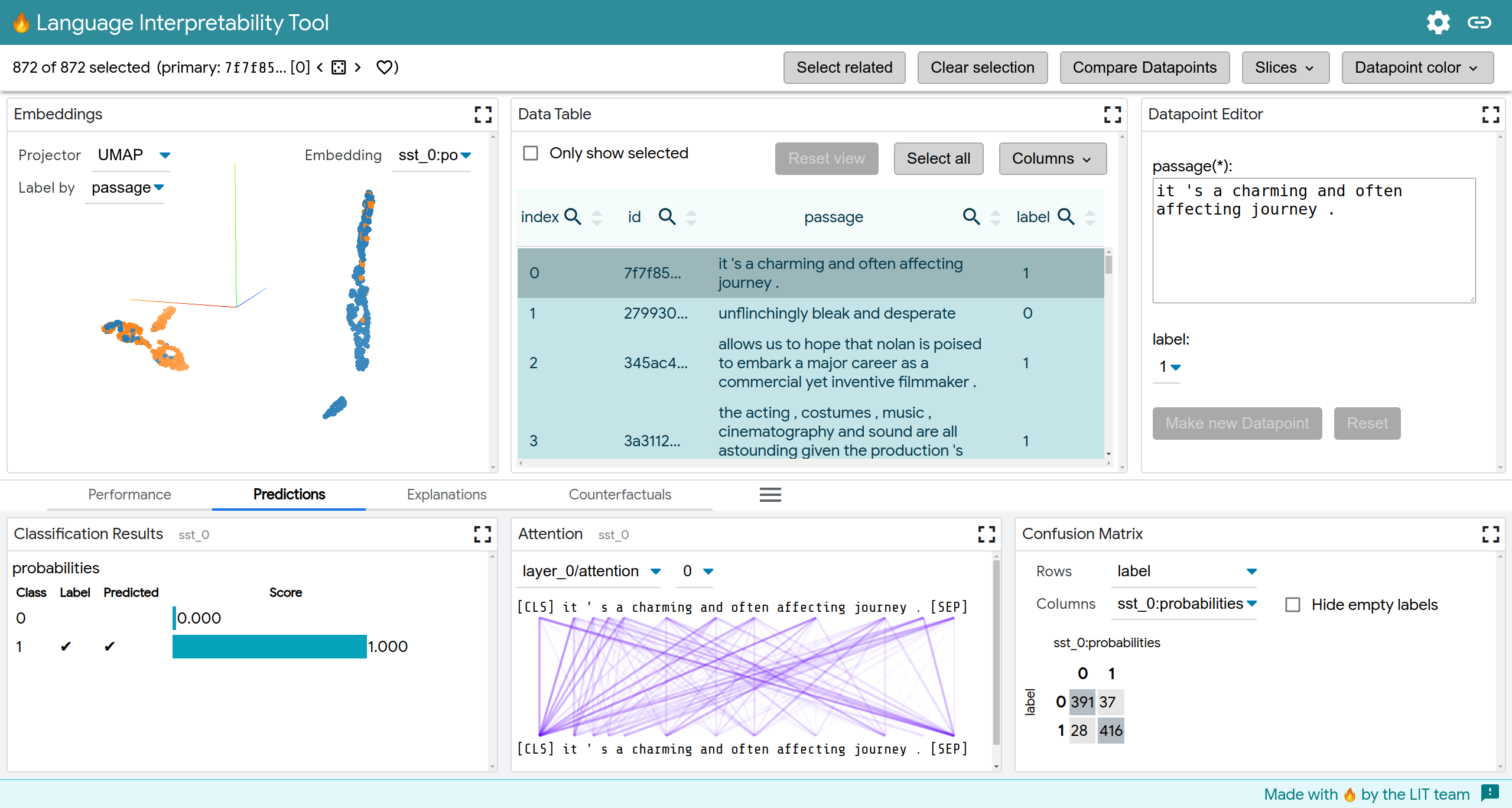}
    \caption{The LIT UI, showing a fine-tuned BERT \citep{DevlinBERT} model on the Stanford Sentiment Treebank \citep{socher2013sst} development set. The top half shows a selection toolbar, and, left-to-right: the embedding projector, the data table, and the datapoint editor. Tabs present different modules in the bottom half; the view above shows classifier predictions, an attention visualization, and a confusion matrix.
    }
    \label{fig:screenshot}
\end{figure*}

With this in mind, we introduce the Language Interpretability Tool (LIT), a toolkit and browser-based user interface (UI) for NLP model understanding. LIT supports local explanations---including salience maps, attention, and rich visualizations of model predictions---as well as aggregate analysis---including metrics, embedding spaces, and flexible slicing---and allows users to seamlessly hop between them to test local hypotheses and validate them over a dataset. LIT provides first-class support for counterfactual generation: new datapoints can be added on the fly, and their effect on the model visualized immediately.
Side-by-side comparison allows for two models, or two datapoints, to be visualized simultaneously. 

We recognize that research workflows are constantly evolving, and designed LIT along the following principles:
\begin{itemize}\itemsep0em
    \item \textbf{Flexible:} Support a wide range of NLP tasks, including classification, seq2seq, language modeling, and structured prediction. 
    \item \textbf{Extensible:} Designed for experimentation, and can be reconfigured and extended for novel workflows.
    \item \textbf{Modular:} Components are self-contained, portable, and simple to implement.
    \item \textbf{Framework agnostic:} Works with any model that can run from Python ---including TensorFlow \citep{tensorflow-paper}, PyTorch \citep{pytorch-paper}, 
    or remote models on a server.
    \item \textbf{Easy to use:} Low barrier to entry, with only a small amount of code needed to add models and data (\autoref{sec:running-lit}), and an easy path to access sophisticated functionality.
\end{itemize}

\section{User Interface and Functionality}
\label{sec:workflows}

\begin{table*}[tbh]  %
  \centering
  \small
    \begin{tabular*}{\textwidth}{p{4cm}p{11cm}}
    \toprule
    \textbf{Module}           & \textbf{Description}\\
    \midrule
    \textbf{Attention} & Displays an attention visualization for each layer and head.\\
    \midrule
    \textbf{Confusion Matrix} & A customizable confusion matrix for single model or multi-model comparison.\\
    \midrule
    \textbf{Counterfactual Generator} & Creates counterfactuals for selected datapoint(s) using a variety of techniques.\\
    \midrule
    \textbf{Data Table}       & A tabular view of the data, with sorting, searching, and filtering support.\\
    \midrule
    \textbf{Datapoint Editor} & Editable details of a selected datapoint.\\
    \midrule
    \textbf{Embeddings}       & Visualizes dataset by layer-wise embeddings, projected down to 3 dimensions.\\
    \midrule
    \textbf{Metrics Table}    & Displays metrics such as accuracy or BLEU score, on the whole dataset and slices.\\
    \midrule
    \textbf{Predictions} & Displays model predictions, including classification, text generation, language model probabilities, and a graph visualization for structured prediction tasks. \\
    \midrule
    \textbf{Salience Maps} & Shows heatmaps for token-based feature attribution for a selected datapoint using techniques like local gradients and LIME.\\
    \midrule
    \textbf{Scalar plot} & Displays a jitter plot organizing datapoints by model output scores, metrics or other scalar values. \\
    \bottomrule
    \end{tabular*}
    \caption{Built-in modules in the Language Interpretability Tool. \label{tab:module_table}}
\end{table*}

LIT has a browser-based UI comprised of modules (\autoref{fig:screenshot}) which contain controls and visualizations for specific tasks (Table \ref{tab:module_table}). 
At the most basic level, LIT works as a simple demo server: one can enter text, press a button, and see the model's predictions. But by loading an evaluation set, allowing dynamic datapoint generation, and an array of interactive visualizations, metrics, and modules that respond to user input, LIT supports a much richer set of user journeys:

\paragraph{J1 - Explore the dataset.} Users can interactively explore datasets using different criteria across modules like the data table and the embeddings module (similar to \citet{smilkov2016embedding}), in which a PCA or UMAP \citep{mcinnes2018umap} projection can be rotated, zoomed, and panned to explore clusters and global structures (\autoref{fig:screenshot}-top left).

\paragraph{J2 - Find interesting datapoints.} Users can identify interesting datapoints for analysis, cycle through them, and save selections for future use. For example, users can select off-diagonal groups from a confusion matrix, examine outlying clusters in embedding space, or select a range based on scalar values (Figure~\ref{fig:s2s} (a)).

\paragraph{J3 - Explain local behavior.} Users can deep-dive into model behavior on selected individual datapoints using a variety of modules depending on the model task and type. For instance, users can compare explanations from salience maps, including local gradients \citep{li-etal-2016-visualizing} and LIME \citep{ribeiro2016lime}, or look for patterns in attention heads (\autoref{fig:screenshot}-bottom).

\paragraph{J4 - Generate new datapoints.} Users can create new datapoints based on datapoints of interest either manually through edits, or with a range of automatic counterfactual generators, such as backtranslation \citep{bannard-callison-burch-2005-paraphrasing}, nearest-neighbor retrieval \citep{andoni2006near}, word substitutions (``great $\rightarrow$ terrible''), or adversarial attacks like HotFlip \citep{ebrahimi2018hotflip} (\autoref{fig:counterfactuals}). Datapoint provenance is tracked to facilitate easy comparison.

\paragraph{J5 - Compare side-by-side.} Users can interactively compare two or more models on the same data, or a single model on two datapoints simultaneously. Visualizations automatically ``replicate'' for a side-by-side view.

\paragraph{J6 - Compute metrics.} LIT calculates and displays metrics for the whole dataset, the current selection, as well as on manual or automatically-generated slices  (\autoref{fig:cs-winogender} (c)) to easily find patterns in model performance.

\bigskip
LIT's interface allows these user journeys to be explored interactively. Selecting a dataset and model(s) will automatically show compatible modules in a multi-pane layout (\autoref{fig:screenshot}). A tabbed bottom panel groups modules by workflow and functionality, while the top panel shows persistent modules for dataset exploration.

These modules respond dynamically to user interactions. If a selection is made in the embedding projector, for example, the metrics table will respond automatically and compute scores on the selected datapoints. Global controls make it easy to page through examples, enter a comparison mode, or save the selection as a named ``slice''.
In this way, the user can quickly explore multiple workflows using different combinations of modules.

A brief video demonstration of the LIT UI is available at \url{https://youtu.be/j0OfBWFUqIE}.

\section{Case Studies}
\label{sec:case-studies}

\begin{figure}[t!]
    \centering
    \fcolorbox{black!20}{white}{
    \includegraphics[width=0.95\columnwidth,clip,trim=3mm 3mm 3mm 4mm]{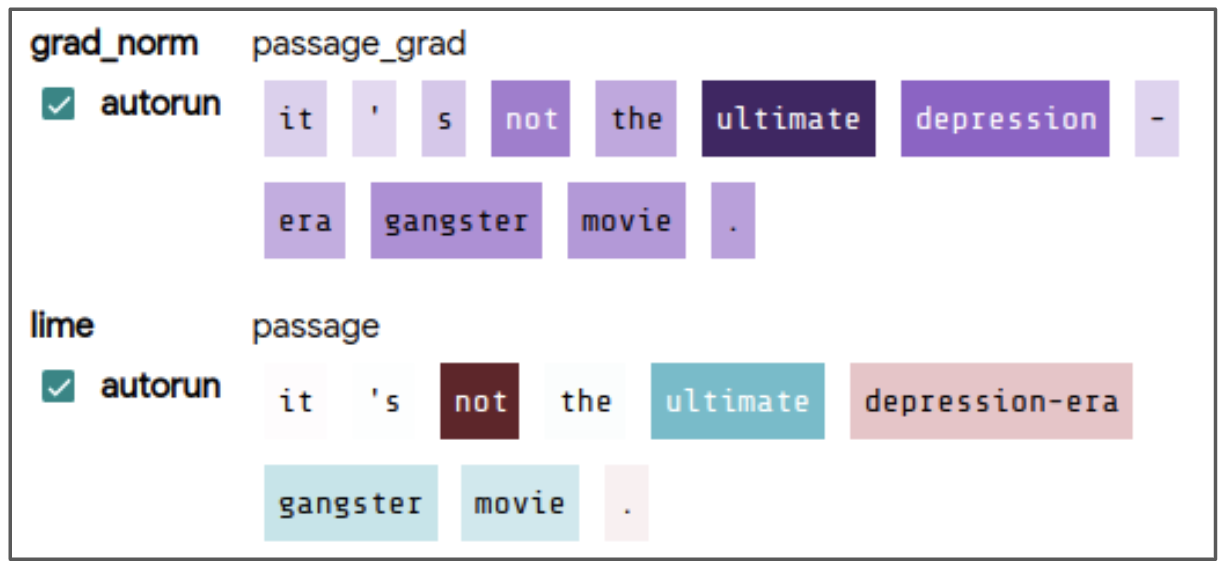}
    }
    \caption{Salience maps on ``\textit{It's not the ultimate depression-era gangster movie.}'', suggesting that ``not'' and ``ultimate'' are important to the model's prediction.}
    \label{fig:cs-sent-analysis}
\end{figure}

\paragraph{Sentiment analysis.} How well does a sentiment classifier handle negation? We load the development set of the Stanford Sentiment Treebank \citep[SST;][]{socher2013sst}, and use the search function in LIT's data table \textbf{(J1, J2)} to find the 56 datapoints containing the word ``not''. Looking at the Metrics Table \textbf{(J6)}, we find that surprisingly, our BERT model \citep{DevlinBERT} gets 100\% of these correct! But we might want to know if this is truly robust. With LIT, we can select individual datapoints and look for explanations \textbf{(J3)}. For example, take the negative review, ``\textit{It's \underline{not} the ultimate depression-era gangster movie.}''. 
As shown in \autoref{fig:cs-sent-analysis}, salience maps suggest that ``not'' and ``ultimate'' are important to the prediction.

We can verify this by creating modified inputs, using LIT's datapoint editor \textbf{(J4)}. Removing ``not'' gets a strongly positive prediction from ``\textit{It's the ultimate depression-era gangster movie.}'', while replacing ``ultimate'' to get ``\textit{It's \underline{not} the \underline{worst} depression-era gangster movie.}'' elicits a mildly positive score from our model.

\begin{figure}[t!]
    \centering
    \includegraphics[width=0.90\columnwidth]{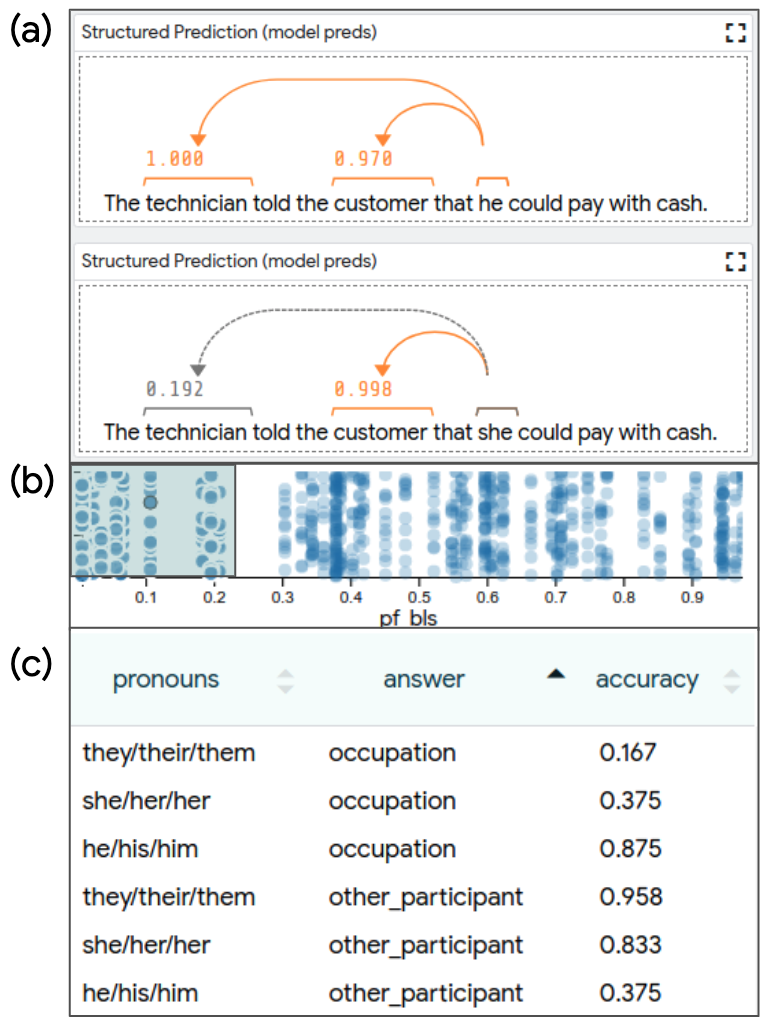}
    \caption{Exploring a coreference model on the Winogender dataset.}
    \label{fig:cs-winogender}
\end{figure}

\paragraph{Gender bias in coreference.} Does a system encode gendered associations, which might lead to incorrect predictions? We load a coreference model trained on OntoNotes \citep{hovy2006ontonotes}, and load the Winogender \citep{rudinger2018winogender} dataset into LIT for evaluation. Each Winogender example has a pronoun 
and two candidate referents, one a occupation term like (``technician'') and one an ``other participant'' (like ``customer''). Our model predicts coreference probabilities for each candidate.
We can explore the model's sensitivity to pronouns by comparing two examples side-by-side (see \autoref{fig:cs-winogender} (a).)
We can see how commonly the model makes similar errors by paging through the dataset (\textbf{J1}), or by selecting specific slices of interest. For example, we can use the scalar plot module (\textbf{J2}) (\autoref{fig:cs-winogender} (b)) to select datapoints where the occupation term is associated with a high proportion of male or female workers, according to the U.S. Bureau of Labor Statistics \citep[BLS;][]{caliskan2017semantics}.

In the Metrics Table \textbf{(J6)}, we can slice this selection by pronoun type and by the true referent. On the set of male-dominated occupations ($<$ 25\% female by BLS), we see the model performs well when the ground-truth agrees with the stereotype - e.g. when the answer is the occupation term, male pronouns are correctly resolved 83\% of the time, compared to female pronouns only 37.5\% of the time (\autoref{fig:cs-winogender} (c)).

\begin{figure}[t!]
    \centering
    \includegraphics[width=\columnwidth]{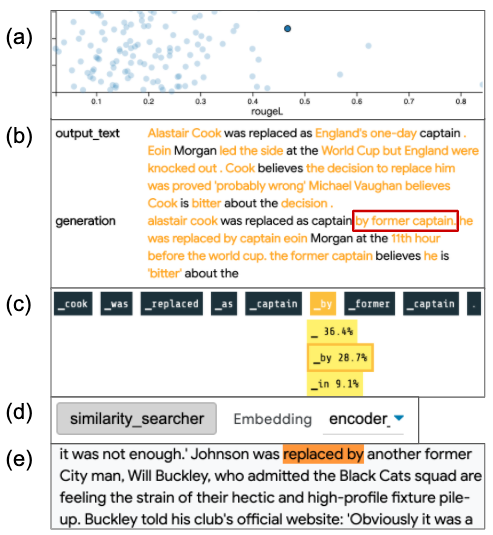}
    \caption{Investigating a local generation error, from selection of an interesting example to finding relevant training datapoints that led to an error.}
    \label{fig:s2s}
\end{figure}

\paragraph{Debugging text generation.} Does the training data explain a particular error in text generation? We analyze a T5 \citep{raffel2019exploring} model on the CNN-DM summarization task \citep{hermann2015teaching}, and loosely follow the steps of \citet{strobelt2018seq2seqvis}. LIT's scalar plot module \textbf{(J2)} allows us to look at per-example ROUGE scores, and quickly select an example with middling performance (\autoref{fig:s2s} (a)). We find the generated text (\autoref{fig:s2s} (b)) contains an erroneous constituent: ``\textit{alastair cook was replaces as captain \underline{by former captain} ...}''. We can dig deeper, using LIT's language modeling module (\autoref{fig:s2s} (c)) to see that the token ``by'' is predicted with high probability (28.7\%).

To find out how T5 arrived at this prediction, we utilize the ``similarity searcher'' component through the counterfactual generator tab (\autoref{fig:s2s} (d)). This performs a fast approximate nearest-neighbor lookup \citep{andoni2006near} from a pre-built index over the training corpus, using embeddings from the T5 decoder. With one click, we can retrieve 25 nearest neighbors and add them to the LIT UI for inspection (as in \autoref{fig:counterfactuals}). We see that the words ``captain'' and ``former'' appear 34 and 16 times in these examples--along with 3 occurrences of ``replaced by'' (\autoref{fig:s2s} (e))--suggesting a strong prior toward our erroneous phrase.

\section{System design and components}
\label{sec:system}

The LIT UI is written in TypeScript, and communicates with a Python backend that hosts models, datasets, counterfactual generators, and other interpretation components. LIT is agnostic to modeling frameworks; data is exchanged using NumPy arrays and JSON, and components are integrated through a declarative ``spec'' system (\autoref{sec:spec}) that minimizes cross-dependencies and encourages modularity. A more detailed design schematic is given in the Appendix, \autoref{fig:design}.

\subsection{Frontend}
\label{sec:frontend}
The browser-based UI is a single-page web app, built with lit-element\footnote{\url{https://lit-element.polymer-project.org/}. Naming is coincidental; the Language Interpretability Tool is not related to the lit-html or lit-element projects.} and MobX\footnote{\url{https://mobx.js.org/}}. A shared framework of ``service'' objects tracks interaction state, such as the active model, dataset, and selection, and coordinates a set of otherwise-independent modules which provide controls and visualizations.

\subsection{Backend}
\label{sec:backend}
The Python backend serves models, data, and interpretation components. The server is stateless, but includes a caching layer for model predictions, which frees components from needing to store intermediate results and allows interactive use of large models like BERT \citep{DevlinBERT} and GPT-2 \citep{radford2019language}. Component types include:
\begin{itemize}\itemsep0em
    \item \textbf{Models} which implement a \texttt{predict()} function, \texttt{input\_spec()}, and \texttt{output\_spec()}.
    \item \textbf{Datasets} which load data from any source and expose an \texttt{.examples} field and a \texttt{spec()}.
    \item \textbf{Interpreters} are called on a model and a set of datapoints, and return output---such as a salience map---that may also depend on the model's predictions.
    \item \textbf{Generators} are interpreters that return new input datapoints from source datapoints.
    \item \textbf{Metrics} are interpreters which return aggregate scores for a list of inputs.
\end{itemize}
These components are designed to be self-contained and interact through minimalist APIs, with most exposing only one or two methods plus spec information. They communicate through standard Python and NumPy types, making LIT compatible with most common modeling frameworks, including TensorFlow \citep{tensorflow-paper} and PyTorch \citep{pytorch-paper}. Components are also portable, and can easily be used in a notebook or standalone script. For example:
\begin{minted}[fontsize=\footnotesize,autogobble,frame=single]{python}
  dataset = SSTData(...)
  model = SentimentModel(...)
  lime = lime_explainer.LIME()
  lime.run([dataset.examples[0]], 
           model, dataset)
\end{minted}
will run the LIME \citep{ribeiro2016lime} component and return a list of tokens and their importance to the model prediction.

\subsection{Running with your own model}
\label{sec:running-lit}
LIT is built as a Python library, and its typical use is to create a short \verb|demo.py| script that loads models and data and passes them to the \verb|lit.Server| class:
\begin{minted}[fontsize=\footnotesize,autogobble,frame=single]{python}
  models = {'foo': FooModel(...),
            'bar': BarModel(...)}
  datasets = {'baz': BazDataset(...)}
  server = lit.Server(models, datasets)
  server.serve()
\end{minted}

A full example script is included in the Appendix (\autoref{fig:demo-py}). The same server can host several models and datasets for side-by-side comparison, and can also reference remotely-hosted models.

\subsection{Extensibility: the \texttt{spec()} system}
\label{sec:spec}
NLP models come in many shapes, with inputs that may involve multiple text segments, additional categorical features, scalars, and more, and output modalities that include classification, regression, text generation, and span labeling. Models may have multiple heads of different types, and may also return additional values like gradients, embeddings, or attention maps. Rather than enumerate all variations, LIT describes each model and dataset with an extensible system of semantic types.

For example, a dataset class for textual entailment \citep[][]{dagan-et-al-2006-textual-entailment,bowman-etal-2015-large}
might have \texttt{spec()}, describing available fields:
\begin{itemize}\itemsep-0.5em
  \item \textbf{premise:} \verb|TextSegment()|
  \item \textbf{hypothesis:} \verb|TextSegment()|
  \item \textbf{label:} \verb|MulticlassLabel(vocab=...)|
\end{itemize}
A model for the same task would have an \texttt{input\_spec()} to describe required inputs:
\begin{itemize}\itemsep-0.5em
  \item \textbf{premise:} \verb|TextSegment()|
  \item \textbf{hypothesis:} \verb|TextSegment()|
\end{itemize}
As well as an \texttt{output\_spec()} to describe its predictions:
\begin{itemize}\itemsep-0.5em
  \item \textbf{probas:} \verb|MulticlassPreds(| \\ \verb|    vocab=..., parent="label")|
\end{itemize}

Other LIT components can read this spec, and infer how to operate on the data. For example, the \verb|MulticlassMetrics| component searches for \verb|MulticlassPreds| fields (which contain probabilities), uses the \verb|vocab| annotation to decode to string labels, and evaluates these against the input field described by \verb|parent|. Frontend modules can detect these fields, and automatically display: for example, the embedding projector will appear if \verb|Embeddings| are available. 

New types can be easily defined: a \verb|SpanLabels| class might represent the output of a named entity recognition model, and custom components can be added to interpret it. 

\section{Related Work}
\label{sec:related-work}

A number of tools exist for interactive analysis of trained ML models. 
Many are general-purpose, such as the What-If Tool \citep{what-if-tool}, Captum \citep{captum2019github}, Manifold \citep{zhang2018manifold}, or InterpretML \citep{nori2019interpretml}, while others focus on specific applications like fairness, including FairVis \citep{cabrera2019fairvis} and FairSight \citep{Ahn_2019}. And some provide rich support for counterfactual analysis, either within-dataset (What-If Tool) or dynamically generated as in DiCE \citep{mothilal2020explaining}. 

For NLP, a number of tools exist for specific model classes, such as RNNs \citep{strobelt2017lstmvis}, Transformers \citep{Hoover_Exbert,Vig_Analyzing}, or text generation \citep{strobelt2018seq2seqvis}. More generally, AllenNLP Interpret \citep{wallace-etal-2019-allennlp} introduces a modular framework for interpretability components, focused on single-datapoint explanations and integrated tightly with the AllenNLP \citep{Gardner2017AllenNLP} framework. 

While many components exist in other tools, LIT aims to integrate local explanations, aggregate analysis, and counterfactual generation into a single tool. In this, it is most similar to Errudite \citep{wu-etal-2019-errudite}, which provides an integrated UI for NLP error analysis, including a custom DSL for text transformations and the ability to evaluate over a corpus. However, LIT is explicitly designed for flexibility: we support a broad range of workflows
and provide a modular design for extension with new tasks, visualizations, and generation techniques.

\paragraph{Limitations} LIT is an evaluation tool, and as such is not directly useful for training-time monitoring.
As LIT is built to be interactive, it does not scale to large datasets as well as offline tools such as TFMA \citep{mewald2018tfma}. (Currently, the LIT UI can handle about 10,000 examples at once.) Because LIT is framework-agnostic, it does not have the deep model integration of tools such as AllenNLP Interpret \citep{wallace-etal-2019-allennlp} or Captum \citep{captum2019github}.
This makes many things simpler and more portable, but also requires more code for techniques like integrated gradients \citep{pmlr-v70-sundararajan17a} that need to directly manipulate parts of the model.

\section{Conclusion and Roadmap}
\label{sec:conclusion}

LIT provides an integrated UI and a suite of components for visualizing and exploring the behavior of NLP models. It enables interactive analysis both at the single-datapoint level and over a whole dataset, with first-class support for counterfactual generation and evaluation. LIT supports a diverse range of workflows, from explaining individual predictions to disaggregated analysis to probing for bias through counterfactuals. LIT also supports a range of model types and techniques out of the box, and is designed for extensibility through simple, framework-agnostic APIs.

LIT is under active development by a small team. Planned upcoming additions include new counterfactual generation plug-ins, additional metrics and visualizations for sequence and structured output types, and a greater ability to customize the UI for different applications. 

LIT is open-source under an Apache 2.0 license, and we welcome contributions from the community at \url{https://github.com/pair-code/lit}.

\bibliographystyle{acl_natbib}
\bibliography{main,data,models,tools}

\appendix
\setcounter{figure}{0} \renewcommand{\thefigure}{A.\arabic{figure}}
\onecolumn
\section{Appendices}
\label{sec:appendix}

\begin{figure*}[h!]
    \centering
    \includegraphics[width=\textwidth]{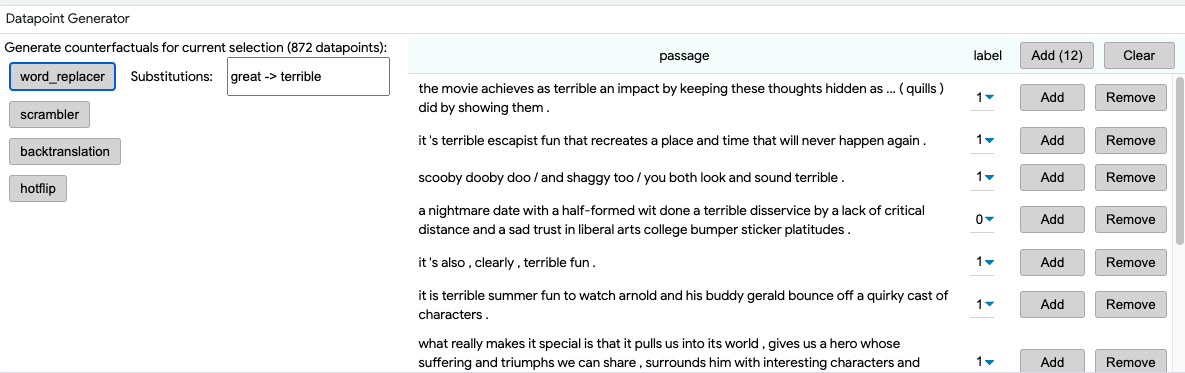}
    \caption{The counterfactual generator module, showing a set of generated datapoints in the staging area. Labels can be maually edited before adding these to the dataset. In this example, the counterfactuals were created using the word replacer, replacing the word ``great'' with ``terrible'' across the dataset.}
    \label{fig:counterfactuals}
\end{figure*}

\begin{figure*}[h!]
    \centering
    \includegraphics[width=\textwidth]{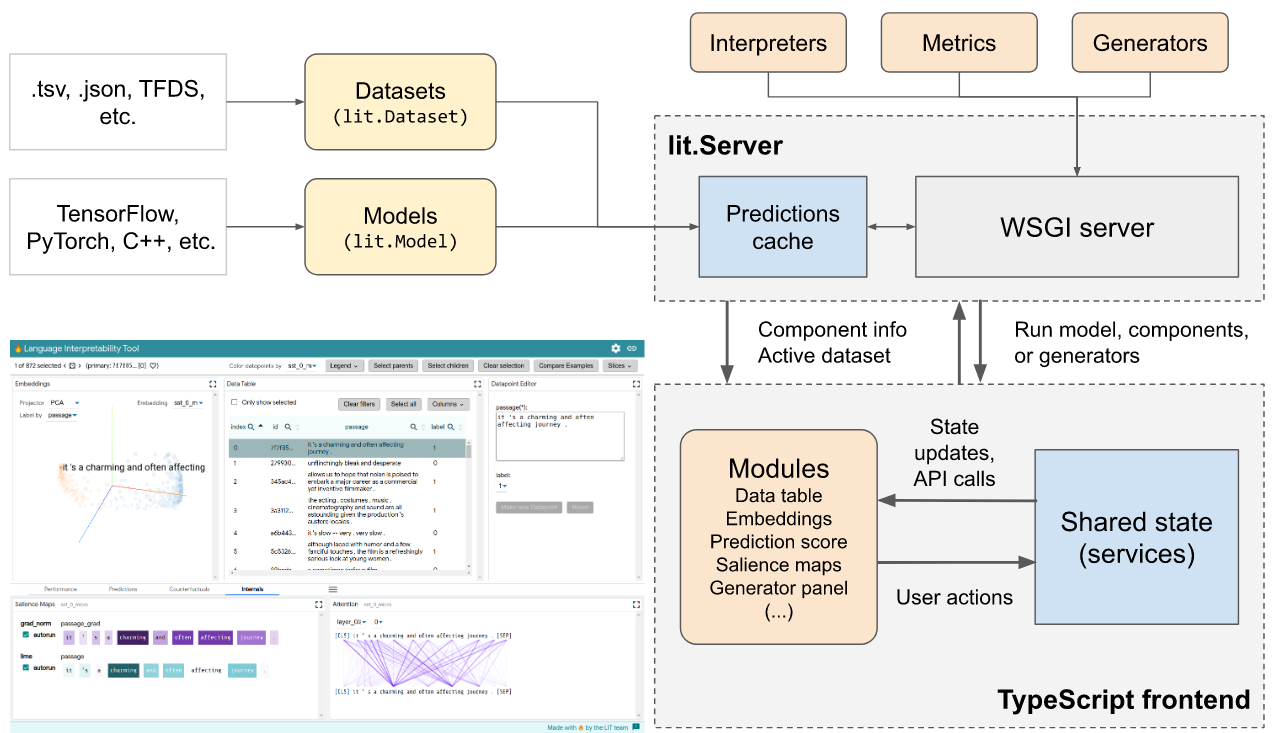}
    \caption{Overview of LIT system architecture. The backend manages models, datasets, metrics, generators, and interpretation components, as well as a caching layer to speed up interactive use. The frontend is a TypeScript single-page app consisting of independent modules (webcomponents built with lit-element) which interact with shared ``services'' which manage interaction state. The backend can be extended by passing components to the \texttt{lit.Server} class in the demo script (\autoref{sec:running-lit} and \autoref{fig:demo-py}), while the frontend can be extended by importing new components in a single file, \texttt{layout.ts}, which both lists available modules and specifies their position in the UI (\autoref{fig:screenshot}).}
    \label{fig:design}
\end{figure*}

\begin{figure*}
    \centering
    \usemintedstyle{vs}
\begin{minted}{python}
NLI_LABELS = ['entailment', 'neutral', 'contradiction']

class MultiNLIData(lit.Dataset):
  """Loader for MultiNLI dataset."""

  def __init__(self, path):
    # Read the eval set from a .tsv file
    df = pandas.read_csv(path, sep='\t')
    # Store as a list of dicts, conforming to self.spec()
    self._examples = [{
      'premise': row['sentence1'],
      'hypothesis': row['sentence2'],
      'label': row['gold_label'],
      'genre': row['genre'],
    } for _, row in df.iterrows()]

  def spec(self):
    return {
      'premise': lit_types.TextSegment(),
      'hypothesis': lit_types.TextSegment(),
      'label': lit_types.Label(vocab=NLI_LABELS),
      # We can include additional fields, which don't have to be used by the model.
      'genre': lit_types.Label(),
    }


class MyNLIModel(lit.Model):
  """Wrapper for a Natural Language Inference model."""

  def __init__(self, model_path, **kw):
    # Load the model into memory so we're ready for interactive use.
    self._model = _load_my_model(model_path, **kw)

  ##
  # LIT API implementations
  def predict(self, inputs: List[Input]) -> Iterable[Preds]:
    """Predict on a single minibatch of examples."""
    examples = [self._model.convert_dict_input(d) for d in inputs]  # any custom preprocessing
    return self._model.predict_examples(examples)  # returns a dict for each input

  def input_spec(self):
    """Describe the inputs to the model."""
    return {
        'premise': lit_types.TextSegment(),
        'hypothesis': lit_types.TextSegment(),
    }

  def output_spec(self):
    """Describe the model outputs."""
    return {
      # The 'parent' keyword tells LIT where to look for gold labels when computing metrics.
      'probas': lit_types.MulticlassPreds(vocab=NLI_LABELS, parent='label'),
      # This model returns two different embeddings, but you can easily add more.
      'output_embs': lit_types.Embeddings(),
      'mean_word_embs':  lit_types.Embeddings(),
      # In LIT, we treat tokens as another model output. There can be more than one,
      # and the 'align' field describes which input segment they correspond to.
      'premise_tokens': lit_types.Tokens(align='premise'),
      'hypothesis_tokens': lit_types.Tokens(align='hypothesis'),
      # Gradients are also returned by the model; 'align' here references a Tokens field.
      'premise_grad': lit_types.TokenGradients(align='premise_tokens'),
      'hypothesis_grad': lit_types.TokenGradients(align='hypothesis_tokens'),
      # Similarly, attention references a token field, but here we want the model's full "internal"
      # tokenization, which might be something like: [START] foo bar baz [SEP] spam eggs [END]
      'tokens': lit_types.Tokens(),
      'attention_layer0': lit_types.AttentionHeads(align=['tokens', 'tokens']),
      'attention_layer1': lit_types.AttentionHeads(align=['tokens', 'tokens']),
      'attention_layer2': lit_types.AttentionHeads(align=['tokens', 'tokens']),
      # ...and so on. Since the spec is just a dictionary of dataclasses, you can populate it
      # in a loop if you have many similar fields.
    }


def main(_):
  datasets = {
      'mnli_matched': MultiNLIData('/path/to/dev_matched.tsv'),
      'mnli_mismatched': MultiNLIData('/path/to/dev_mismatched.tsv'),
  }

  models = {
      'model_foo': MyNLIModel('/path/to/model/foo/files'),
      'model_bar': MyNLIModel('/path/to/model/bar/files'),
  }

  lit_demo = lit.Server(models, datasets, port=4321)
  lit_demo.serve()

if __name__ == '__main__':
  main()
\end{minted}
    \caption{Example demo script to run LIT with two NLI models and the MultiNLI \citep{Williams_MNLI} development sets. The actual model can be implemented in TensorFlow, PyTorch, C++, a REST API, or anything that can be wrapped in a Python class: to work with LIT, users needs only to define the spec fields and implement a \texttt{predict()} function which returns a dict of NumPy arrays for each input datapoint. The dataset loader is even simpler; a complete implementation is given above to read from a TSV file, but libraries like TensorFlow Datasets can also be used.}
    \label{fig:demo-py}
\end{figure*}

\begin{figure*}[h!]
    \centering
    \includegraphics[width=\textwidth]{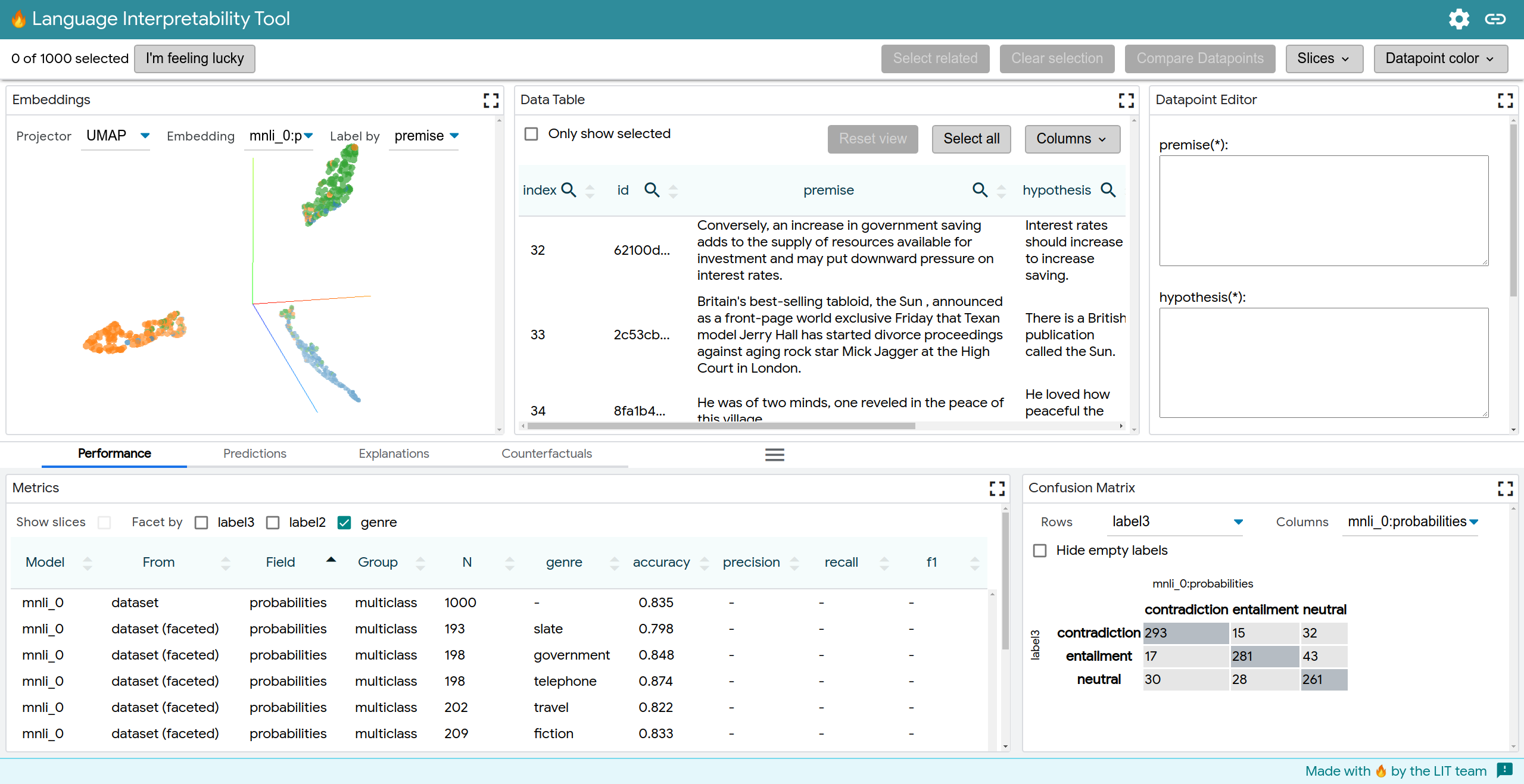}
    \caption{Full UI screenshot, showing a BERT \citep{DevlinBERT} model on a sample from the ``matched'' split of the MultiNLI \citep{Williams_MNLI} development set. The embedding projector (top left) shows three clusters, corresponding to the output layer of the model, and colored by the true label. On the bottom, the metrics table shows accuracy scores faceted by genre, and a confusion matrix shows the model predictions against the gold labels.}
    \label{fig:example-mnli-metrics}
\end{figure*}

\begin{figure*}[h!]
    \centering
    \begin{subfigure}{0.3\textwidth}
        \centering
        \includegraphics[width=\linewidth]{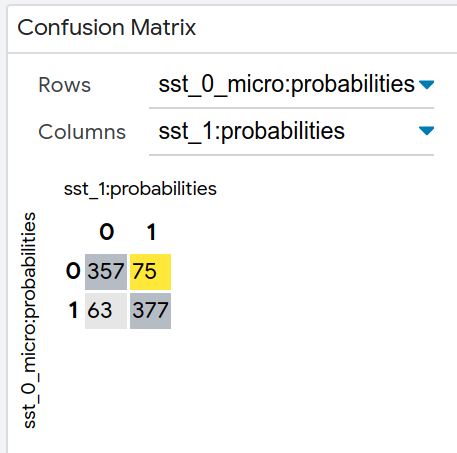}
        \caption{}
    \end{subfigure}
    \begin{subfigure}{\textwidth}
        \centering
        \includegraphics[width=\linewidth]{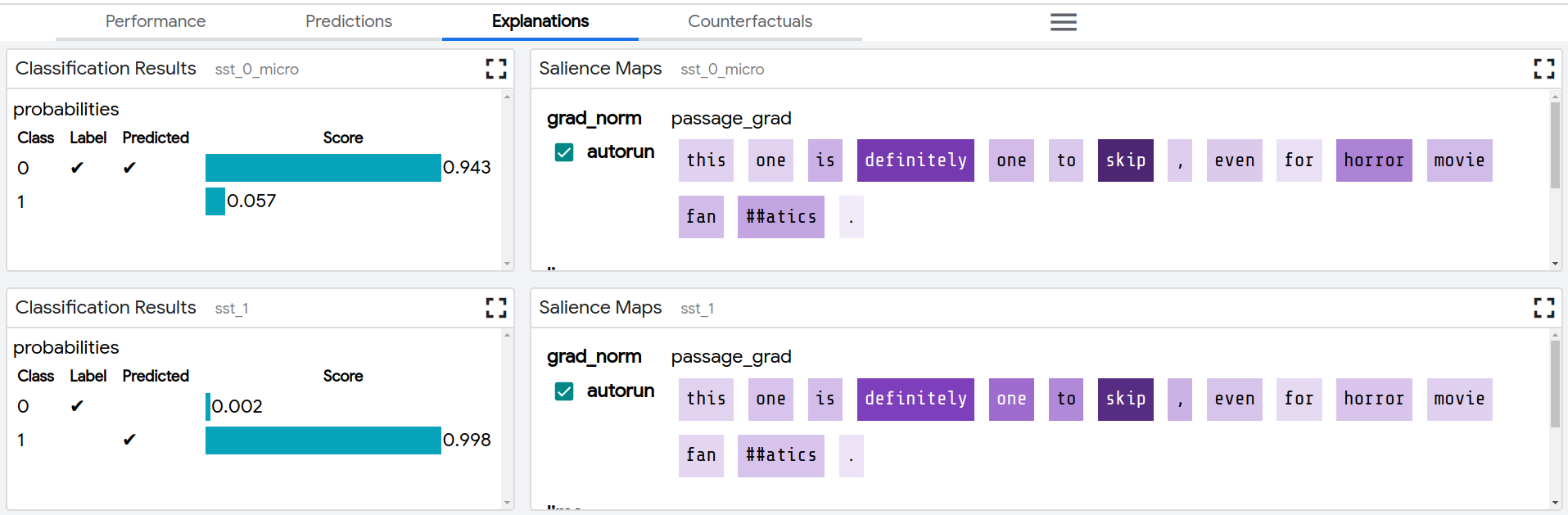}
        \caption{}
    \end{subfigure}
    \caption{Confusion matrix (a) and side-by-side comparison of predictions and salience maps (b) on two sentiment classifiers. In model comparison mode, the confusion matrix can compare two models, and clicking an off-diagonal cell with select examples where the two models make different predictions. In (b) we see one such example, where the model in the second row (``sst\_1'') predicts incorrectly, even though gradient-based salience show both models focusing on the same tokens.}
    \label{fig:example-sst-comparison}
\end{figure*}

\end{document}